\documentclass[10pt,a4paper]{scrartcl}
\usepackage[top=0.85in,left=2in,footskip=0.75in,marginparwidth=2in]{geometry}

\usepackage[utf8]{inputenc}
\usepackage{graphicx}
\usepackage{wrapfig}

\usepackage{color}
\definecolor{Gray}{gray}{.25}

\usepackage{url}
\usepackage{amsmath,amssymb}
\usepackage{natbib}
\usepackage{enumerate}

\usepackage[aboveskip=1pt,labelfont=bf,labelsep=period,singlelinecheck=off,font=small]{caption}
\usepackage{sidecap}
\usepackage[T1]{fontenc}
\usepackage{times}
\usepackage{multirow}

\usepackage{tikz}
\usepackage{forest}
\usepackage{subcaption}
\usepackage[colorinlistoftodos]{todonotes}

\title{The Logoscope: a Semi-Automatic Tool for Detecting and Documenting French New Words}
\subtitle{From the Linguistic Project to the Web Interface}
\author{Ingrid Falk \and Delphine Bernhard \and Christophe Gérard}

\begin{document}

\maketitle              

\begin{abstract}
  In this article we present the design and implementation of the
  \textit{Logoscope}, the first tool especially developed to
  detect new words of the French language, to document them and allow
  a public access through a web interface.  This semi-automatic tool collects new words daily by browsing the online versions of French
  well known newspapers such as Le Monde, Le Figaro, L’Équipe,
  Libération, La Croix, Les Échos.  In contrast to other existing
  tools essentially dedicated to dictionary development, the \textit{Logoscope} attempts to give a more complete account of the
   context in which the new words occur.  In addition
  to the commonly given morpho-syntactic information it also provides information about the textual and
  discursive contexts of the word creation; in particular, it
  automatically determines the (journalistic) topics of the text
  containing the new word.
In this article we first give a general overview of the developed tool.  
We then describe the approach
  taken, we discuss the linguistic background which guided our design
  decisions and present the computational methods we used to implement
  it.


\end{abstract}

\tableofcontents
\newpage
\section{Introduction: New Words, NLP and the Logoscope}
\label{sec:introduction}
%

%

Like other areas of linguistics, the study of new words 
has benefited from the development of natural language processing techniques in the last twenty years. In this research area, the digital revolution has significantly changed the way to collect the data necessary for empirical research \citep{mejri_presentation:_2011,humbley_neologie_2016}: The traditional practice of collecting new words \textit{by reading} texts (newspapers, literature, scientific and technical texts, etc.),\footnote{For the French language, the manual detection method has produced the \textit{Base d'Observation et de Recherche des Néologismes}, \url{http://www.atilf.fr/borneo}, ATILF - CNRS \& Nancy Université.} has been supplemented by an automatisation of the collection process. In fact, today there are many computer tools capable of searching through large amounts of texts (often newspapers) in order to automatically detect the newly created words in various languages (German, Catalan, Spanish, English, French, etc.).

\paragraph{What is a new word?}
is, however, one of the first questions to ask when addressing this topic. This is a very difficult question, for which, to our knowledge there is no clear consensual answer in the scientific community.

A naïve but plausible standpoint is to consider new words to be all the lexical units which are absent from all the lexicons that were developed in the history of a language. Besides being hardly feasible, a reason for which this approach is problematic is that arguably these newly created words are not really present or important in the contemporary ``textual public use''. Often they are hapaxes which were used in one text only (\textit{eg.}, \textit{quadrambulation}, \textit{sucraphone}, etc.\footnote{Theses hapaxes were created by \textit{newologism}(@NewThingFriends) on Twitter: ``Quadrambulation: Walking around the block'' (June 5, 2016), ``Sucraphone: A speaker of sweet nothings'' (July 7, 2016).}) or in few texts (\textit{eg.}, \textit{bredictable}). In other cases their social diffusion is currently too low to be able to speak of collective \textit{use} (\textit{eg.}, \textit{artwashing}). Other word creations, like \textit{brexit}, clearly reflect a societal actuality, are widely publicised but therefore clearly not new. 
On the other hand some words which have already been integrated in a general dictionary (as \textit{phubbing} and \textit{hot take} which were included in this year's Oxford Dictionary) or in a specialised glossary (terminology) could arguably be considered as new by a typical newspaper reader.

From the standpoint of the \textit{Logoscope} ideally both hapaxes and words like \textit{brexit} should be monitored because their distribution in public texts over time provides insights about the life cycle of new words and about the interaction between texts in different domains of the public space and lexical innovation. 

Computational natural language processing methods can easily collect from online texts those forms which are not in a predefined list of known words. However, when using this approach one would encounter several obstacles. Firstly, it is not feasible to compile all the known words of a language (from all dictionaries or terminologies) into such a predefined list. Secondly, online texts are noisy and computational methods are never perfect, so the extracted unknown forms would surely not always be valid words of the considered language. For these reasons we consider that a manual validation will always be necessary. But even a linguist expert can not possibly know all the valid words of a language and there is no generally accepted definition of what a new word is. Therefore ultimately the decision whether a word is new is necessarily subjective, the linguist expert has to rely on the context and on her knowledge and intuition.  

Based on these considerations, our tool was designed to first identify unknown forms using a necessarily incomplete list of known words. Then a linguist expert decides which of these are interesting word creations, based on lexicographic criteria but also taking into consideration the societal and cultural context in which the form occured.

In this article we adopt a pragmatic position and talk of ``new words'', ``recent words'' or ``recently / newly created words'' rather than ``neologism'' and ``neology'', ``lexical creation'', ``neonym'', ``constructed word'', ``lexical innovation'', etc. Indeed, this raises well known terminology issues \citep{guilbert_les_1975,rey_neologisme_1976,boulanger_sur_2010,sablayrolles_terminologie_2006,pruvost_les_2012,cabre_castellvi_neologia:_2015,cabre_per_2016}  that we cannot discuss here, especially since this would require an extensive reflection on the meta-language of linguistics \citep{neveu_pour_2008,swiggers_terminologie_2010}, which is well beyond the scope of this work. However, for ease of use we will sometimes employ the word ``neologism'', but exclusively as a synonym to ``new word''.

These thousands of new words detected each year may be used for several purposes.
They are essential for making general language dictionaries, for driving language policies (Quebec, Spain, etc.), for the observation of science and technology, or to measure the vitality of languages and dialects \citep{cabre_neologia_2000}. Furthermore, lexical creation constantly testifies, within a language area, of economic and political events, of the formation of social identities (slang, etc.), or changes in attitudes (\textit{eg.}, \citep{borde_tirons_2016}), and more broadly, cultural history \citep{calvet_mediterranee:_2016}. Finally, most new words express a rhetorical dimension that is crucial in political communication, advertising or creative writing.

However, to use the collection of new words to this end it is necessary to dispose of a tool addressed not only to 
lexicographers and terminologists, but also to other user communities (journalism, translation, marketing, social sciences, etc.). For these users the appearance of new words may also be relevant since it correlates with events of interest for their particular community (technical, societal, political, sportive, etc.). This is precisely the particular ambition of the framework presented in this paper, the \textit{Logoscope}, and what distinguishes it from other existing IT tools. To this end the new word creations are documented not only using well-known lexicographic categories describing the form itself, but also by taking into account their historical and co(n)textual environment.
In the \textit{Logoscope} framework we achieved this by first developing a principled semi-automatic method to set up a collection of higher-level, coarse-grained topics reflecting what journalistic articles are generally about. This way we could describe the newspaper articles in terms of the topics they were handling and in the same time assess the thematic context in which the word creation in a particular article occurred.


The article is organised as follows. In Section \ref{sec:framework_description} we give a general overview of the \textit{Logoscope} framework. Section \ref{sec:previous-work} is an account of previous work on the automatic detection and documentation of new words, while Section ~\ref{sec:project-background} focuses on the linguistic issues. Section~\ref{sec:nlp-techniques} details the NLP methods and techniques implemented in the \textit{Logoscope}. Finally, Section \ref{sec:application} presents some sample findings obtained by using this tool.


\section{General Description of the \textit{Logoscope} Framework}
\label{sec:framework_description}

The main building block of the \textit{Logoscope} framework consists of a continuously updated lexical resource containing unknown words, their occurrences in French online press articles and further morpho-syntactic and contextual documentation. 
In addition our system features a query interface which allows an open access to this knowledge base.\footnote{A simplified online version is available at the url \url{http://logoscope.unistra.fr}. See also Section~\ref{sec:web-interface} where we give some more details.} 

\subsection{New Words are Collected in Three Stages}
\label{sec:system-overview}


\begin{figure}[h!]
\includegraphics[scale=0.58]{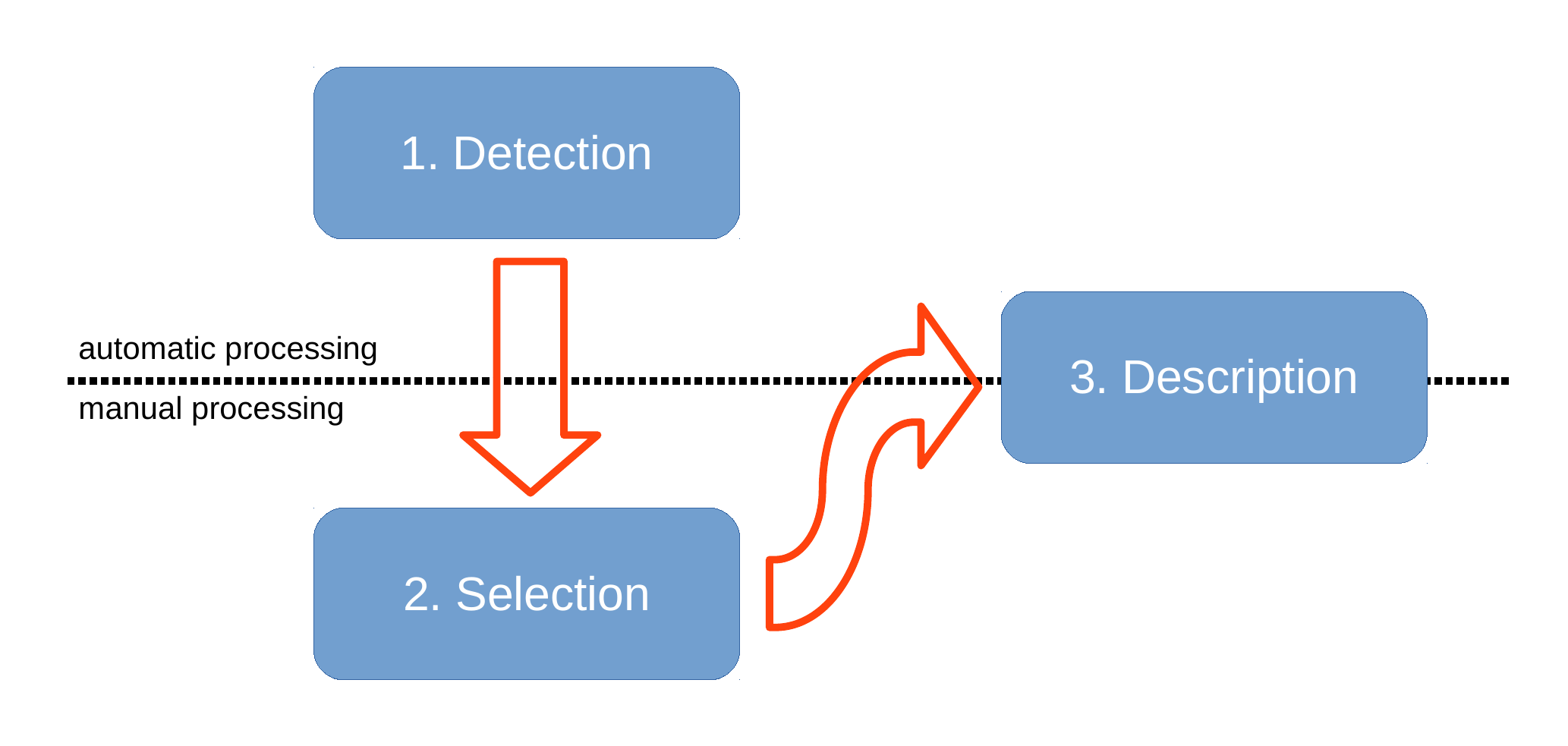}
\caption{Processing stages in the \textit{Logoscope}.\label{fig:stages}}
\end{figure}

The \textit{Logoscope} is built in three stages (see Figure \ref{fig:stages}). The first stage consists in the automatic detection of the unknown words. In the second stage the detected unknown words are manually validated, that is, a human expert decides which of them are genuine word creations. Finally, in the third stage these validated unknown words are documented. This is done firstly by manually adding morpho-syntactic features and information on the originating creation process. This information is automatically complemented by contextual information, typically \textit{e.g} the journal and the paragraph the new words appeared in. But, in contrast to other similar systems (see Section~\ref{sec:syst-neol-detect}), the \textit{Logoscope} also (automatically) gives a rough approximation of what the article containing the new word is about.

In the next sections we will first give a more detailed description of each of these three stages before briefly addressing how the \textit{Logoscope} might be used through its web interface in Section~\ref{sec:web-interface}. A more detailed report of the arising NLP challenges and the methods we used to tackle them will be presented in Section~\ref{sec:nlp-techniques}.

\subsection{Stage 1: Detection of Unknown Words}
\label{sec:detect-unkn-words}

The \textit{Logoscope} retrieves newspaper articles from several RSS feeds\footnote{Currently we collect articles from the following newspapers: \textit{Le Monde}, \textit{Libération}, \textit{L'Équipe}, \textit{Le Figaro}, \textit{Les Echos}, \textit{La Croix}.} in French on a daily basis\footnote{The average number of sources collected per day is 430 with a maximum and minimum of 968 and 28 respectively.}. 
The newspaper articles are preprocessed such that only the journalistic content is kept. The articles are then segmented into paragraphs and word forms. The resulting forms are filtered based on an exclusion list (French words found in several lexicons and corpora). 
They are then reordered in such a way that those words which are the most likely new word candidates appear on top, using a supervised classification method which will be described more in detail in Section~\ref{sec:detect-unkn-words-impl}.

\subsection{Stage 2: Selection of the Valid Candidates}
\label{sec:valid-detect-unkn}

\begin{figure}
  \centering
  \begin{subfigure}[b]{0.55\textwidth}
    \scalebox{0.8}{
      \noindent\begin{tabular}{@{}|lllll|}
        \hline
        wid,&word,&valid,&pos,&proc,\\
        \hline
        4800,&lumbersexuel,&0,&,&,\\
        6333,&syllogistique,&0,&,&,\\
        7802,&ressurgissant,&0,&,&,\\
        8068,&fantasmatiquement,&0,&,&,\\
        4953,&franco-planétaire,&0,&,&,\\
        6214,&consommatoire,&0,&,&,\\
        8974,&kill-billeuse,&0,&,&,\\
        4232,&normcore,&0,&,&,\\
        5401,&sex-pics,&0,&,&,\\
        \hline
      \end{tabular}
    }
    \subcaption{Sample csv (comma separated values) file listing the detected unknown words on 2015-01-02 before validation.\label{fig:ex-unknown-csv}}
  \end{subfigure}
  \quad
  \begin{subfigure}[b]{0.40\textwidth}
    \includegraphics[width=\textwidth]{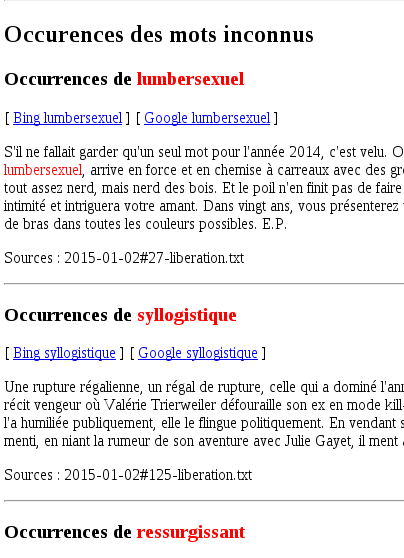}
    \subcaption{Contexts of unknown words (as html, viewed in a browser).\label{fig:ex-unknown-html}}
  \end{subfigure}
  \caption{Sample list of automatically detected unknown words and their contexts as presented to the linguist expert for validation.\label{fig:sample-unknown-words-validation}}
\end{figure}
The list of unknown words produced in the previous stage is presented to a linguist expert together with the context they appeared in, for validation. Figure~\ref{fig:sample-unknown-words-validation} shows the first entries of a list generated this way (Fig.~\ref{fig:ex-unknown-csv}) and the corresponding contexts (Fig.~\ref{fig:ex-unknown-html}). The list is completed by a linguist expert who adds the following information:
\begin{enumerate}[a.]
\item\label{item:neo} whether the form is a genuine new word or not (column 3), \textit{e.g.} ``syllogistique'' is not a new word.
\item\label{item:gramcat} the grammatical category of the new word, \textit{e.g.} ADJ (adjective), NOM (noun) (column 4),
\item\label{item:proc} the type of creation process which engendered the new word (column 5), \textit{e.g.} MORSEM (morphosemantic), EMP (borrowing).\footnote{To add this information the linguist expert uses a classification introduced by \cite{sablayrolles_jean-francois_neologia_2011}. In Section~\ref{sec:project-background} we will explain the linguistic background and motivation for this decision.}
\end{enumerate}
The first decision (\ref{item:neo}) is the most difficult and time
consuming. For the simplest cases it can be made based on the context (the surrounding paragraph)
which is presented together with the unknown form (as shown in
Figure~\ref{fig:ex-unknown-html}), but often it also involves looking
up dictionaries or the original article or performing additional investigations on the Web. It is at this point and based on this process that the linguist expert determines which new words are to enter the system and be further documented.

A typical result of this process is shown in Figure~\ref{fig:ex-unknown-csv-validated}. 

\begin{figure}
  \centering
    \noindent\begin{tabular}{@{}|lllll|}
      \hline
      wid,&word,&valid,&pos,&proc,\\
      \hline
      4800,&lumbersexuel,&1,&NOM,&MORSEM,\\
      6333,&syllogistique,&0,&,&,\\
      7802,&ressurgissant,&0,&,&,\\
      8068,&fantasmatiquement,&0,&,&,\\
      4953,&franco-planétaire,&1,&ADJ,&MORSEM,\\
      6214,&consommatoire,&1,&ADJ,&MORSEM,\\
      8974,&kill-billeuse,&1,&NOM,&MORSEM,\\
      4232,&normcore,&1,&NOM,&EMP,\\
      5401,&sex-pics,&0,&,&,\\
      \multicolumn{5}{|c|}{...}\\
      \hline
    \end{tabular}
    \caption{Sample csv (comma separated values) file showing the result of the validation process for the unknown words detected on 2015-01-02.}
    \label{fig:ex-unknown-csv-validated}	
\end{figure}

\subsection{Stage 3: Linguistic Description of the Selected New Words} 
\label{sec:docum}

In the next stage, the validated new words are documented, that is they are added to our knowledge base together with the following pieces of information:
\begin{enumerate}[a.]
\item\label{item:pos-proc} the grammatical category and the type of the creation process (added by the linguist expert in the previous stage, as described in Section~\ref{sec:valid-detect-unkn}),
\item\label{item:context} the paragraph containing the new word,
\item\label{item:topics} further contextual information about the containing article:
  \begin{enumerate}[i.]
  \item\label{item:journal} the journal,
  \item the publication date,
  \item the author (if available),
  \item\label{item:position} the position in the text (beginning, middle or end of the article),
  \item\label{item:them-anal} the thematic context, represented by the three most prominent general journalistic topics addressed in the article.
  \end{enumerate}
\end{enumerate}

While the information in (\ref{item:pos-proc}) has been added manually during the validation stage, the surrounding paragraph (\ref{item:context}) and the contextual information regarding the containing article (\ref{item:topics}) are added automatically. Whereas adding most of these items (\ref{item:context}, \ref{item:journal}-\ref{item:position}) is straightforward, automatically detecting the topics of an article requires a thematic analysis which, to our knowledge, is not proposed by any other similar tool. We will describe the methods used for its implementation in Section~\ref{sec:thematic-analysis}.

\begin{figure}
  \centering
  \includegraphics[width=\textwidth]{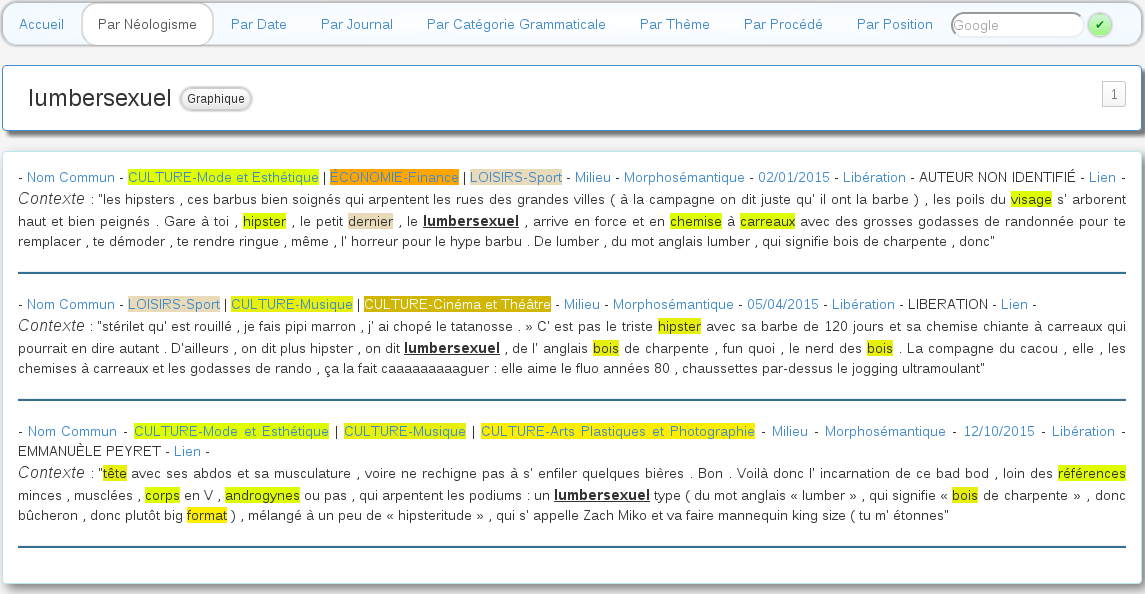}
  \caption{Documentation of the new word \textit{lumbersexuel}. 
  }
  \label{fig:ex-lumbersexuel}
\end{figure}

After the selection of a new word and its linguistic description, this new form is further monitored, that is we check the collected sources for subsequent occurrences and describe them in the same way as for the first occurrence. The result of this process is shown in Figure~\ref{fig:ex-lumbersexuel} for the new word \textit{lumbersexuel}.\footnote{Borrowed from the English \textit{lumbersexual}, ``A male hipster who affects a rugged, outdoorsy look, typified by plaid shirts and a full beard. '' (Wiktionary, retrieved on 08-07-2016))} The user can see that this word first appeared in January 2015 in the journal \textit{Libération}, that it is a common noun and that its production involved a morphosemantic process. It was then automatically detected again in April and October 2015, again in \textit{Libération} and each time it occurred in a paragraph situated in the middle of the containing article. For each occurrence the system gives the surrounding paragraph. 

As regards point (\ref{item:them-anal}), the most prominent themes in the containing article are presented in different colours, as well as the characteristic words of these themes occurring in the context. In this case the result of the thematic analysis was that the articles containing \textit{lumbersexuel} are mostly related to culture and leisure (\textit{i.e.} they contain many terms related to culture and leisure). In Section \ref{sec:project-background} we will explain in detail why we attach particular importance to these themes.
Finally the exclusion list is updated with all manually checked candidates.

\subsection{The Web Interface: Using the \textit{Logoscope}}
\label{sec:web-interface}

Since the platform is targeted both at non specialists and more expert users
from various scientific domains (\textit{e.g.} linguistics and lexicography but
also social sciences, politics or economics) the Web interface is
designed to allow queries of a varying degree of complexity.

The most straightforward access to the knowledge base is provided by
a Web interface (\url{http:://logoscope.unistra.fr/}) where the
new words can be browsed by different criteria, as for example by
grammatical category, date, journal, etc. (see Figure~\ref{fig:ex-lumbersexuel}).

\begin{figure}[htb]
  \centering
  \includegraphics[width=\textwidth]{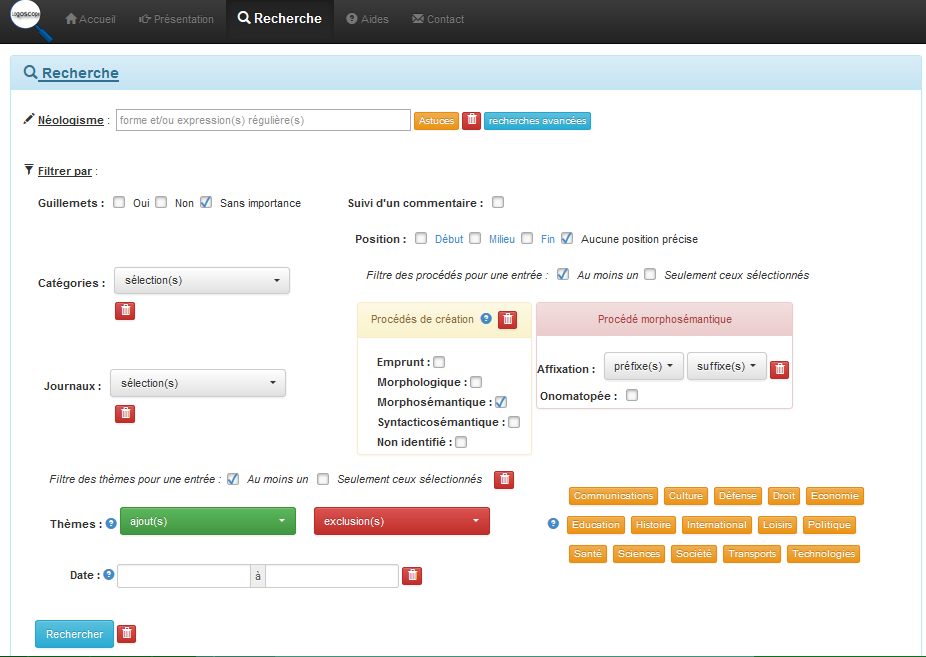}
  \caption{Query page of the \textit{Logoscope} browser-based application.}
  \label{fig:web-interface-query}
\end{figure}

In addition we also provide a browser-based application which allows more
complex queries. Figure~\ref{fig:web-interface-query} shows a screenshot of this query interface. 
Here a user can use regular expressions to search for
new words or query the knowledge base by several criteria (for example find
new words between quotation marks, in a given period of time, occurring
in articles with a given predominant topic or a combination thereof). 

\section{Previous Work}
\label{sec:previous-work}

%
%

In this section we will first present other systems devoted to the detection and documentation of new words (Section~\ref{sec:syst-neol-detect})
before addressing previous work related to the two components of the \textit{Logoscope} which mainly rely on automatic methods.
%
The first of these two components, the automatic detection of new words, is presented  in Section~\ref{sec:detect-unkn-words-pw}. 
In Section~\ref{sec:docum-foll} we describe work related to the documentation of the detected words, focusing on an outstanding feature of the \textit{Logoscope}, namely an approach to the automatic thematic analysis of their context. 

\subsection{Existing Systems and Resources for Detecting and Documenting New Words}
\label{sec:syst-neol-detect}

The approaches used by systems concerned with the semi-automatic detection and observation of word creation in different languages
can be analysed according to different criteria. Some of these are:
\begin{enumerate}
\item\label{item:constitution} How the new words are detected: manually, automatically, etc.
\item\label{item:sources} The sources where new words are extracted from. These can be for example various online resources, but also documents provided by the user.
\item\label{item:documentation} The information associated with the new word and presented to the end user. For most systems the goal is to elaborate a lexicographic resource. Therefore they acquire lexical information in the first place but also more or less elaborate data about the context in which the word creation occurred.
\end{enumerate}

The simplest systems are databases, constituted manually and intuitively over some period of time from a collection of available documents.
A typical representative of this approach is \textsc{Borneo}, a database of French neologisms.\footnote{\url{http://web.atilf.fr/BORNEO-Base-d-Observation-et-de}}
Another group of systems, \textit{e.g.} \textsc{Pompamo} \citep{ollinger_creativite_2010}, are pure extraction utilities which are fed with a text in which unknown words are detected and analysed.

In the third more elaborate category of frameworks, the systems consist of (1) a new word detection stage based on a methodologically well defined dynamic corpus followed by (2) a documentation phase where the detected and documented new words are added to 
the actual linguistic knowledge base. Prominent examples of this type of approach are \textsc{Obneo} \citep{Cabre_et_Al_2003,estopa_r._metodologia_2004}, a framework for Spanish and Catalan and \textsc{Neologia} (for French, \citealt{sablayrolles_neologia_2010,sablayrolles_jean-francois_neologia_2011,cartier_neologie_2011}). 

These systems also differ with respect to how the new words are detected. Whereas for \textsc{Wortwarte}\footnote{\url{http://www.wortwarte.de}} \citep{lothar_lemnitzer_neologismenlexikographie_2010}, a German neologism detection system, the acquisition process is almost entirely automated and requires little human intervention\footnote{One reason for this is, of course, that very few parameters are documented.}, for \textsc{Neologia} the whole documentation process is manual. 

Considering lexical documentation, 
the amount and type of data which are retrieved and documented vary largely. 
Thus, for example the \textsc{Wortwarte}, only records the gender, the plural mark and an ad-hoc lexical domain (\textit{e.g.} education, health, telecommunication, environment, etc.). In contrast, \textsc{Obneo} and even more so \textsc{Neologia} provide a very detailed lexical documentation covering (among others) the following aspects: morpho-syntax (which part of speech, grammatical sub-category \textit{e.g. qualitative adjective}, number, gender), semantics (predicate or argument, hyper-class \textit{e.g. human, animal, etc.}), semantic relations (synonymy, antonomy, etc.). 
In addition \textsc{Neologia} also provides a definition or gloss of the described entry and thus entirely assumes the role of lexicographer.

As already mentioned, the \textit{Logoscope} belongs to the latter type of frameworks, based on both a new word detection and a documentation phase. In the following we will first introduce previous work concerning the detection of new words (Section~\ref{sec:detect-unkn-words-pw}). In Section~\ref{sec:docum-foll} we give a theoretical analysis of the information used for the documentation of new words by the various systems, describe the items we chose for the \textit{Logoscope} and motivate these design choices.

\subsection{Methods for the Detection of New Words}
\label{sec:detect-unkn-words-pw}
%
%

Current methods for the automatic or semi-automatic identification of neologisms  mainly target the coinage of new words. Some recent works have addressed semantic neologisms, with a focus on changes in part-of-speech \citep{janssen_neotag:_2012} or restricted case studies \citep{boussidan_using_2011,reutenauer_vers_2012}.
We do not address these phenomena for the time being.




For the detection of new words, two different types of methods may be distinguished:
\begin{itemize}
\item Methods based on lists containing known words in the target language, which are used to identify unknown words. These lists are usually called \emph{exclusion lists};
\item Methods relying on various statistical measures or machine learning applied to diachronic corpora.
\end{itemize}
The use of exclusion lists is by far the most common method. 
For French, the \textsc{Pompamo} tool \citep{ollinger_creativite_2010} uses an exclusion list made of the French lexicon \textsc{Morphalou 2.0} \citep{romary:inria-00100195}, a list of named entities and lexicons provided by the user. In addition to the lexicons, the tool uses filters which detect non-alphanumeric characters, numbers and composed word forms. \cite{issac_cyberneologisme_2011} also describes several filters, aimed at eliminating unwanted neologism candidates not found in the exclusion list.
The first filter eliminates non words by looking for bigrams and trigrams of characters which are not found in French. The second filter targets words which are concatenated due to missing spaces and looks for all the possible combinations in a dictionary. Finally, spelling errors are identified by finding corrections with the Levenshtein distance. 

Systems relying on exclusion lists have also been developed for languages other than French. \textsc{Cenit} --~Corpus-based English Neologism Identifier Tool~-- \citep{roche_cenit_1999} uses additional filters which aim at detecting proper nouns. For German, the \textsc{Wortwarte} platform\footnote{\url{http://www.wortwarte.de}} \citep{lemnitzer_mots_2012} is an example of this exclusion list based approach.

All these methods rely on simple heuristics and require that the candidates be manually validated by an expert. 




The \textit{Logoscope} attempts to combine the two main approaches. We use exclusion lists as filters but also apply machine learning techniques to select the most probable neologism candidates. Since we expect the users of our system to be interested in the appearance of new words for different reasons and to have different views on the phenomenon, we did not rely on a very specific definition of neologisms. 

\subsection{Features Used for the Description of New Words}
\label{sec:docum-foll}

As already mentioned in Section~\ref{sec:syst-neol-detect}, most complex neology observation systems, namely \textsc{Obneo}, \textsc{Neologia} and \textsc{Wortwarte}, document mainly lexical information. However, in most cases they also add some information related to the context in which the new word occurred. 
This is not surprising since the context of occurrences is particularly important for the documentation of the new words. It not only proves and exemplifies the existence of the new word but also illustrates its meaning, especially for systems which do not provide other explanations or glosses (as is the case for the \textit{Logoscope}).
In addition to lexical specifications and the context, neology systems also provide various other information, as for example:\footnote{For a more in depth discussion please see \cite{logo-cmlf2014}. }
\begin{description}
\item[Appearance date]
\item[Source:] spoken/written, media name, competence domain\footnote{\textsc{Obneo} records the competence domain of the source in its corpus of spontaneous writings (\textit{e.g.} written documents other than newspapers). Surprisingly, by ``type of text'' (tipus de text in Catalan) \textsc{Obneo} does not mean a particular kind of text (\textit{e.g.} law text, scientific review) but a set of publications related either to a particular professional domain (\textit{e.g.} economy, sports, culture, computer science) or to publications of general interest (\textit{i.e.} without a particular domain). \textsc{Neologia}'s ``domain'' documents the sources using similar labels: economy, sport, society, politics, culture, etc.}
\item[Authorship:] name of the author, name of the publisher, etc.\footnote{\textsc{Obneo} also monitors a corpus of radio and television programs (Catalunya Ràdio, COM Ràdio, Televisió de Catalunya, Ràdio 9). To comply with the specificities of this corpus of spoken language, \textsc{Obneo} not only documents parameters intrinsic to spoken corpora like the phonetic transcription but also the role of the speaker (\textit{e.g.} presenter, reporter, etc.), her age and gender, dialect and mother tongue.}
\item[Typography:] dashes, quotation marks, italics, etc.
\end{description}
In addition, each of the more complex neology analysis systems stands out by the documentation of a particular feature: Thus \textsc{Neologia} is the only system specifying the position in the text, \textsc{Obneo} draws the users' attention to the discourse genres and \textsc{Wortwarte} proposes a basic thematic classification of the new words by manually assigning a lexical domain. In the \textit{Logoscope} we chose to take a different approach to the thematic documentation of new words, namely by performing a thematic analysis of the surrounding text. This represents a local view on the thematic description of the new word. It allows to assess to some extent the thematic context in which the new word appeared and can be more easily implemented using automatic methods.

\subsubsection{Thematic Analysis}
\label{sec:topic-modeling}

In Natural Language Processing themes or topics\footnote{In the following we use the terms \textit{theme} and \textit{topic} interchangeably.} are generally represented in a  simplified and pragmatic way, namely as a list of characteristic terms.
While a single term is in most cases ambiguous, a list of such terms allows a reciprocal disambiguation and in consequence the definition of a topic.

Thematic analysis may have several goals, which we discuss in the following:
\begin{enumerate}[1.]
\item\label{item:topicalisation} Detecting topics in a corpus of documents in order to determine what it is about.
\item\label{item:annotation} Thematic annotation of documents based on a set of predefined topics.
\end{enumerate}

\paragraph{(\ref{item:topicalisation}) Detecting topics in a collection of documents.}
A technique often used for the automatic thematic analysis of a collection of documents is Topic Modeling. Topic models are probabilistic graphical models allowing to infer the topics of a collection of documents without the use of any prior knowledge or external resource.  
Several approaches of this type have been proposed, as for example Latent Dirichlet Allocation (LDA, \citealt{blei_topic_2009}). 
These models take a corpus of texts as input and identify a set of ``latent'' topics in the form of lists of words (from the corpus) characterising each theme. The model assigns to each document a finite number of topics and each topic has a probability weight. Likewise the corpus is assigned a fixed number of topics with different probability weights. However the inferred topics are not labelled and therefore an important analysis effort is necessary to name them. It is this kind of approach we used for the thematic analysis in the \textit{Logoscope}.

\paragraph{(\ref{item:annotation}) Thematic annotation of documents.}
When a thematic resource is available, documents can be annotated with respect to this resource. In this case the granularity of the thematic analysis depends on the granularity of the resource used for the annotation. \cite{beust:hal-00256152} proposes a thematic markup tool called ThemeEditor.\footnote{\url{https://beust.users.greyc.fr/ThemeEd}} This utility characterises each theme or topic using manually defined lists of terms. In a text given as input, the terms belonging to each theme are then highlighted using different colours. 
The words which characterize a topic are similar to the keywords or key expressions corresponding to the most important subjects of a document. While in many studies keywords or expressions are extracted based on purely statistic methods, \cite{Medelyan:2008:DAK:1364846.1364848} introduce a method for automatic indexation where the keywords assigned to the documents are controlled by aligning them to an existing thesaurus. Medelyan and Witten first identify which terms from the thesaurus occur in the text. Then they use a supervised model to select the most characteristic terms. The model is trained using various features: the \texttt{tf.idf}\footnote{Term Frequency Inverse Document Frequency} score, the position of the first occurence of a term in the text, the number of words making up the term, semantic relation to other candidate terms in the document. Being supervised, a set of mannually indexed documents are also needed. This number has been estimated to 50--100, which reportedly only required a limited effort.

\paragraph{\textit{Logoscope}'s thematic analysis}
For the thematic analysis implemented in the \textit{Logoscope} we first automatically retrieved a set of general journalistic topics using topic modeling (as in \ref{item:topicalisation}). These were then manually labeled and remodeled into a resource (collection of themes, which in turn are represented by lists of terms) which could then be used to thematically mark up the articles containing the new words (based on methods similar to \ref{item:annotation}).

\section{Linguistic Principles of the Logoscope}
\label{sec:project-background}

In this section we discuss the linguistic motivations which guided our choices concerning the documentation of the new French words. 
We specify which data we decided to add to the \textit{Logoscope} knowledge base for each new word and explain the reasons for these decisions. 

In general, a neology observation system\footnote{Cf. the systems described in Section~\ref{sec:syst-neol-detect}, as for example \textsc{Wortwarte}, \textsc{Obneo} and \textsc{Neologia}.} should not only document a large number of lexical features (\textit{e.g.} grammatical category, verb valency, production type, etc.) but also contextual parameters (\textit{e.g.} appearance date, context, quotation marks, etc.) to characterise the words (cf. \citealt{logo-cmlf2014}).


\paragraph{Lexical features: a minimalistic description.}
In the \textit{Logoscope} we deliberately restricted the number of lexical features to the two which we considered essential, namely the grammatical category and the production process. This minimalistic approach is justified by the following considerations.
First, the \textit{Logoscope} is not meant to offer a lexicographic interpretation: we deliberately do not propose any word definition.
Second, this lexical information is nonetheless sufficient as a basis for further investigation by the user according to their particular centres of interest (journalistic, linguistic, sociological, etc., see Section \ref{sec:application}).
Last but not least, in this way the validation time is minimised, which is crucial considering the relatively large number of daily incurring forms needing validation.\footnote{Currently the system collects $\approx$50 forms on average, after an initial ``burn-in'' period of several weeks where it daily collected more than 100 forms.}

Regarding the grammatical category, we were concerned with choosing a small number of well known part of speech tags but which should ideally cover all potential newly created words. This resulted in a set of currently ten grammatical categories: adjective, adverb, gerund, interjection, common noun, proper noun, past participle, present participle, pronoun, verb.

To best document the second lexical feature, the type of process responsible for the word creation, we follow a categorisation proposed in \cite{sablayrolles_jean-francois_neologia_2011} which is outlined in Table~\ref{tab:LogoLexVar}. Ultimately we use the meta category, the first column in Table~\ref{tab:LogoLexVar} to describe the new words ignoring the more specific description of the creation procedure shown in the second column.%
\begin{table}
  \centering
  \begin{tabular}{|l|l|}
    \hline
    \textbf{(Meta-)category} &  \textbf{Type of lexical creation process} \\
    \hline
    \multirow{5}{*}{morpho-semantic} & affixation \\
    & inflection \\
    & composition(s) \\
    & onomatopea \\
    & paronymy or approximate homonymy \\ \hline
    \multirow{4}{*}{syntactico-semantic} & conversion(s) \\
    & lexical and syntactic combinations \\
    & metaphor \\
    & metonymy \\ \hline
    \multirow{2}{*}{morphologic} & truncation \\
    & abbreviation or using only initials \\ \hline
    \multirow{2}{*}{loan} & identical borrowing \\
    & assimilation of loan words \\ \hline
  \end{tabular}
  \caption{Lexical categories documented by \textit{The Logoscope}.}
  \label{tab:LogoLexVar}
\end{table}
Thus for example, \textit{songwriteuse}\footnote{« L’histoire officielle raconte que Courtney Barnett a 26 ans et qu’elle fut serveuse à
Melbourne. Pêchue, acidulée, tonique, c’est surtout la \textbf{songwriteuse} de ce printemps »
(Libération, 23/03/2015).} will simply be described as a \textit{common noun} and \textit{loan}+\textit{morpho-semantic}, but not as an \textit{assimilation of loan word}+\textit{affixation}.
This way the annotator may gain time and minimise the error rate but nonetheless provide relevant information allowing a subsequent more thorough lexicographic research.

\paragraph{Contextual parameters: a maximalistic description.}
One of the specificities of the \textit{Logoscope} is the automatic
documentation of contextual parameters.  This approach stems from
the fact that each lexical creation is dependent on the
communicational situation it occurs in and is tightly interlinked with
the text in which it is read and understood. Moreover, not only the sense of a new word, but also the associated linguistic and extra-linguistic characteristics, its rhetoric function and even the generated mental representation strongly depend on the particular circumstance in which the interpretative activity happens (cf. \citealt{gerard:halshs-01093234,dal-namer}).
%

Our approach belongs in the larger field of text linguistics which places texts (rather than sentences) at the centre of the study of communication systems (cf. \citealt{gerard:halshs-01093209,sole_2002}). This research direction has been strongly advanced from the 1980s on in particular by researchers in Germanic linguistics (reflected for example in \citealt{peschel_zum_2002}). At present it is notably pursued in studies about particular textual genres, such as novels, blogs, etc. (\textit{e.g.} \citealt{siebold_wort-genre-text:_2000,gerard__2016}).

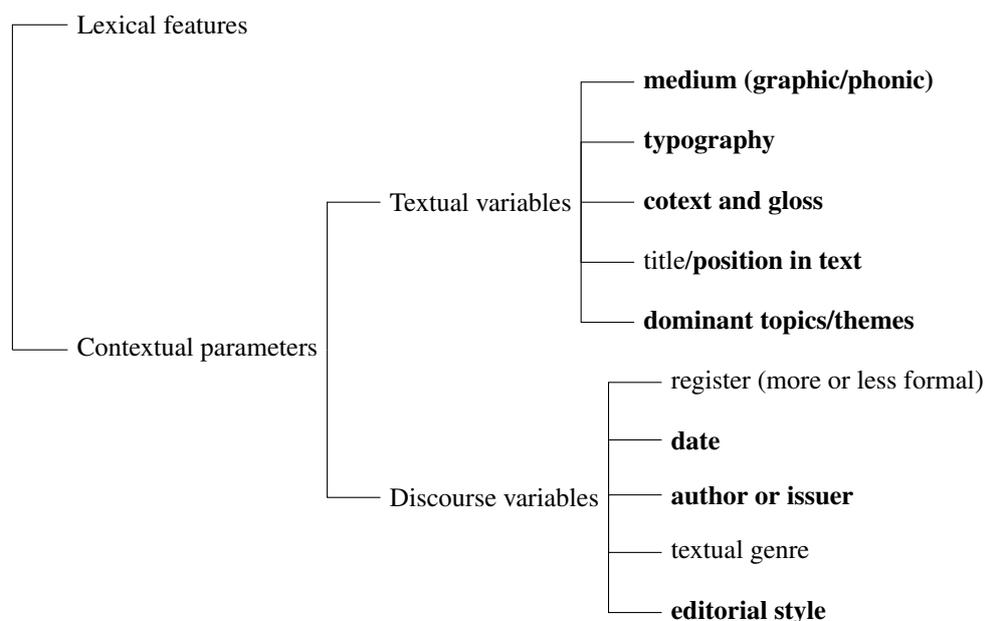
\begin{figure}
  \centering
\begin{forest}
for tree={grow'=0,
anchor=west, child anchor=west, fit=band, parent anchor=east, edge path={\noexpand\path[\forestoption{edge}](!u.parent anchor)|-(.child anchor)\forestoption{edge label};}, l sep=0.7cm,}
[, calign=child,  calign child=2
  [ Lexical features ]
  [ {Contextual parameters} 
     [ {Textual variables} 
        [ {\textbf{medium (graphic/phonic)}}]
        [ {\textbf{typography}}]
        [ {\textbf{cotext and gloss}}]
        [ {title/\textbf{position in text}}]
        [ {\textbf{dominant topics/themes}}]
      ]
      [ {Discourse variables}
        [ {register (more or less formal)}]
        [ {\textbf{date}}]
        [ {\textbf{author or issuer}}]
        [ {textual genre}]
        [ {\textbf{editorial style}}]
      ]
  ]
]
\end{forest}
  
  \caption{Contextual parameters for the description of new words. The parameters set in boldface are those documented in the \textit{Logoscope} framework.}
  \label{fig:neo_vars}
\end{figure}

From a more practical point of view, for our description purpose
we consider that there are two kinds of contextual parameters, namely
textual and discursive variables. As shown in  Figure~\ref{fig:neo_vars}, the former are intrinsically tied to
the semiotic material of the text while the latter rather reflect the
communication situation and the currently valid socio-discursive
norms. In addition to this typology the figure also indicates in
boldface those features which are effectively collected and stored in
the \textit{Logoscope} knowledge base. 

Only the title, the
register and the journalistic genre (\textit{e.g.} editorial, opinion, portrait, interview, etc.) cannot be documented automatically. The
reasons for this are very diverse. While they are of purely
technical nature for the \textit{title} -- the title is marked up in
many different ways, even for the same journal or site -- they are
more fundamental for the automatic detection of the
\textit{journalistic genre}\footnote{To our knowledge \textit{OBNEO} is the only system documenting, manually, the feature of discourse genre by specifying the genre for the radio or television broadcasts in their corpus of spoken language.} and \textit{register} or \textit{code}\footnote{Some systems annotate new words as  ``spoken'' or ``written''. Thus in \textit{BORNEO} words as \textit{craignoss}, \textit{trip musique}, \textit{à chier} or \textit{méga-sale} are assigned an ``oral'' quality. However, we consider that the spoken respectively written quality is not only engendered by the different media but must be viewed in the larger communicational context: The speakers realise that their communicational behaviour must be appropriate to the present setting and they adapt to the situation by adopting a ``style'' which is rather free, less formal or affective or on the contrary more formal, academic or reserved (\citealt{wulf_oesterreicher_gesprochene_2001}).}.  
To our knowledge
currently there are no computational methods sufficiently
sophisticated to allow for the automatic detection of these
variables.

\begin{figure}
  \centering
  \includegraphics[width=\textwidth]{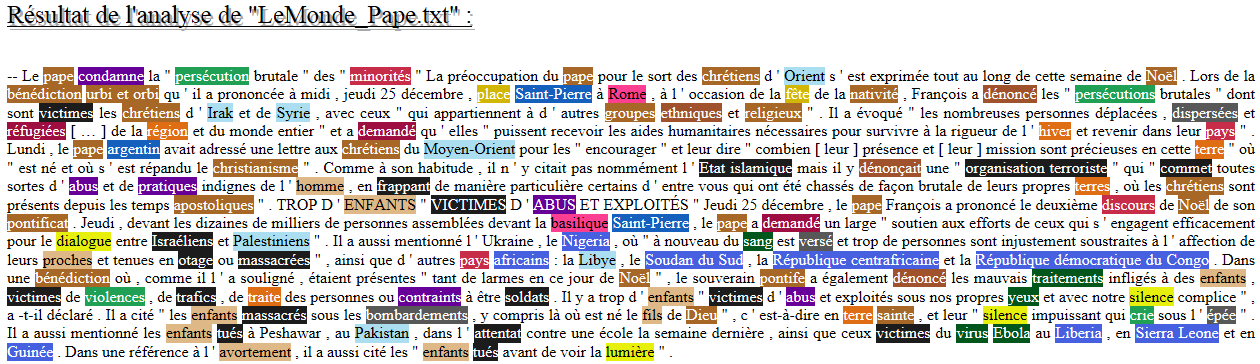}
  \includegraphics[width=\textwidth]{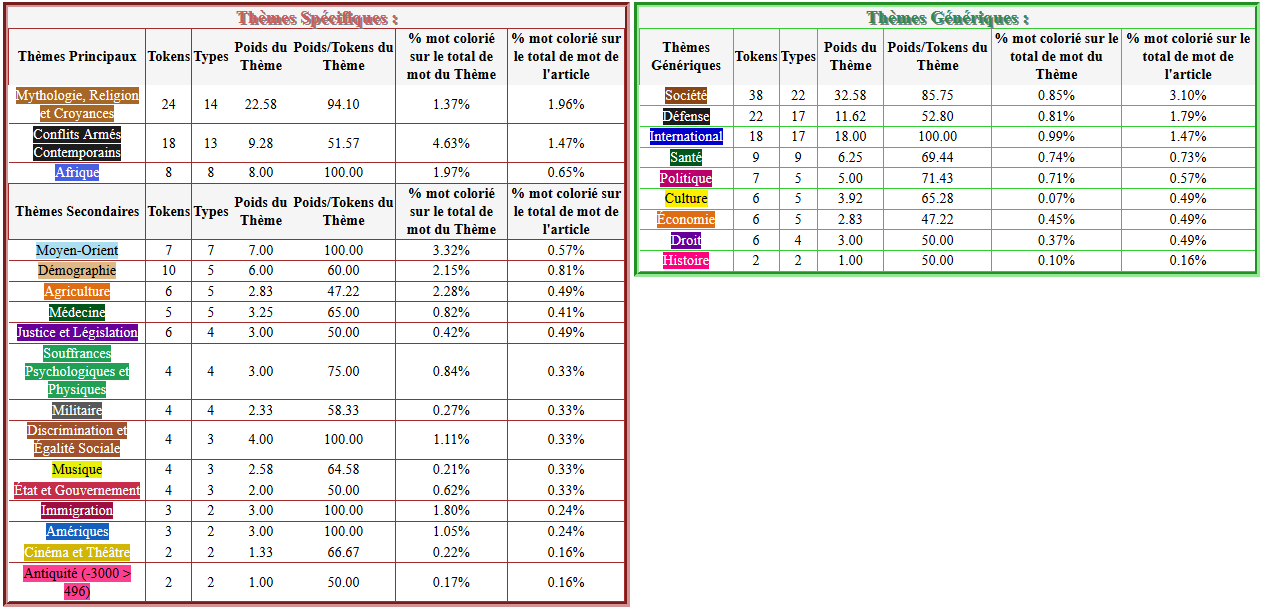}
  \caption{Example of a thematic analysis on a \textit{Le Monde} article (25-12-2015). According to this analysis, the text is principally about a societal topic, but it also employs many words related to the theme of defense in an international context. Within the main topic, the most important subject was found to be about ``mythologies, religion and believes''. In our view, this analysis gives a reasonably good idea about the overall subject of the text, even if it is far from showing anything about the exact content.  }
  \label{fig:ThemAnal-ex}
\end{figure}

\paragraph{The thematic analysis.}
We finish this section by a brief introduction of \textit{Thematic}, a
tool which is dedicated to the thematic analysis of newspaper articles
and thus allows the automatic detection of their dominant themes. The results of this thematic analysis are used to fill in the thematic component of the contextual variables.
They can also be visualised through a thematic colouring as shown in Figure~\ref{fig:ThemAnal-ex}. 

Figure~\ref{fig:ex-lumbersexuel} illustrates how these results are used in the documentation of the new words. It shows that in most cases the articles containing the new word \textit{lumbersexuel} were related to the topics of culture and leisure (green/brown colour).

We will give more details about the theoretical and technical background in Section~\ref{sec:thematic-analysis}.


\section[Implementation]{Implementation: NLP Techniques Employed}
\label{sec:nlp-techniques}

This section describes prominent NLP methods we used in the \textit{Logoscope} and shows how they were implemented to achieve the largely automatic detection and documentation of the new words.
The first set of such NLP techniques, presented in Section~\ref{sec:detect-unkn-words-impl}, is used in the detection of the new words. Section~\ref{sec:thematic-analysis} is about our implementation of a thematic analysis which permits a novel approach to the documentation of the textual context of new words. 

\subsection{Detection of Unknown Words}
\label{sec:detect-unkn-words-impl}

As discussed in
Section~\ref{sec:detect-unkn-words}, most methods addressing the detection of formal neologisms
are based on exclusion lists.  The \textit{Logoscope} also relies on
exclusion lists but in addition the (necessary) manual validation is
alleviated using a statistical machine learning method which ranks the words not in the exclusion list with respect to the probability that they are 
interesting neology candidates.
Here we present a brief overview of our approach, as it already has
been described in detail in \cite{falk:lrec2014}.

As mentioned in Section~\ref{sec:detect-unkn-words}, the \textit{Logoscope} retrieves newspaper articles from several RSS feeds in French on a daily basis. Using exclusion lists it identifies unknown words, which are then presented to a linguist to decide which of these are valid new words.
For example, Table~\ref{tab:unknown-words} shows the most frequent unknown words\footnote{collected on July 12, 2013} resulting from this procedure. The table also illustrates a major drawback of this method. Clearly most of these forms are not interesting neologism candidates: in many cases they are not even valid words and a linguist expert would have to tediously scan a large part of the list before finding interesting candidates.
\begin{table}[htb]
  \centering
  \begin{tabular}{ccc}
lmd (18)&twitter/widgets (7)&india-mahdavi (3)\\
pic(this (18)&garde-à (6)&kilomètresc (2)\\
lazy-retina (9)&ex-PPR (4)&geniculatus (2)\\
onload (9)&pro-Morsi (4)&margin-bottom (2)\\
onerror (9)&tuparkan (4)&politique» (2)\\
amp;euro (7)&candiudature (3)& \ldots\\
  \end{tabular}
  \caption{The most frequent unknown words collected on 2013-07-12. Word frequency is shown in parentheses.}
  \label{tab:unknown-words}
\end{table}
The method we developed ranks these detected unknown forms with respect to the probability that they are interesting neology candidates. Thus, improbable candidates as \textit{e.g. candiudature} or \textit{lmd} would be removed to the bottom of the list whereas more probable candidates as \textit{geniculatus}\footnote{\textit{geniculatus} was not found to be a new word, but it is still a more probable candidate as most other forms in this list} are put at the top of the list.

Our approach is a classical machine learning (supervised classification) approach: We use the extracted forms (and various characteristics or features thereof) and the annotations (valid new word or not) by a linguist expert. Based on this data the system learns a model such that each extracted form is assigned the probability of its being a valid new word. The word forms are then ranked according to this probability. Thus, hopefully, in the validation step, the linguist expert will be presented with the most probable valid new words first.
 
In our approach we experimented with three types of features: form related, morpho-lexical and thematic or contextual features. 
Our experiments highlighted the importance of the thematic features, which, to our knowledge 
have not yet been used in this kind of application. In addition, these features represent a way to access and document the semantic context of the new words.


\paragraph{Features.} We explored the effect on the classification of three groups of features: \textit{formal} features, \textit{morpho-lexical} features and \textit{thematic} features, an overview of which is presented in Table~\ref{tab:features}. The formal features are the most obvious features to be used in such a classification task. They are related to the form or construction of the string at hand, and are language independent. Table~\ref{features:form} shows some examples of such features.
Table~\ref{features:lex} gives an overview of the main morpho-lexical features. 
First, these features check whether particular prefixes and suffixes are present and whether some characters indicate particular languages\footnotemark\footnotetext{We used the \texttt{Lingua::Identify} perl script to this end: \url{http://search.cpan.org/~ambs/Lingua-Identify-0.56/lib/Lingua/Identify.pm}.}.
We also assess the probability that the form might be a spelling error by using the aspell tool\footnotemark\footnotetext{\url{http://aspell.net/}}.
Finally, based on the observation that unknown forms often arise from missing white space we use a further group of morpho-lexical features to check whether other known forms are possibly included in the form at hand\footnote{This group of features is derived from the results of the \cite{Aho:1975:ESM:360825.360855}
string matching algorithm  which suggests a list of known forms present in the unknown form at hand}. 
Obviously these features depend on morpho-lexical characteristics of French.

\begin{table}%
  \begin{subtable}[b]{\textwidth}
    \centering
    \begin{tabular}{|l|}
      \hline
      Length: Number of characters\\
      \hline
      Whether the form contains particular signs, digits, whether it is capitalised,\\
      \hline
      Relative and absolute frequency wrt. to documents and sentences\\
      \hline
    \end{tabular}
    \subcaption{Examples of form related features.\label{features:form}}
  \end{subtable}
  \begin{subtable}[b]{\textwidth}
    \centering
    \begin{tabular}{|ll|}
      \hline
      \multicolumn{2}{|c|}{\textbf{Language}}\\
      \multicolumn{2}{|p{11cm}|}{Whether the form contains characters indicating a particular language (French, English, German or Spanish, 5 features).}\\
      \hline
      \multicolumn{2}{|c|}{\textbf{Prefix/suffix}}\\
      \multicolumn{2}{|p{11cm}|}{0 or 1 depending on whether a particular prefix or suffix is present.}\\
      Prefixes:& ultra-, super-, dé-, ré-, \ldots (69 in total) \\
      Suffixes:& -iste, -ation, -isme, -itude, \ldots (30 in total)\\
      \hline
      \multicolumn{2}{|c|}{\textbf{Spelling}}\\
      \multicolumn{2}{|p{11cm}|}{Levenshtein distance to suggestions from spell-checker (\textit{aspell}) or very large value} \\
      \multicolumn{2}{|p{11cm}|}{Does the form contain other known forms? (\citealt{Aho:1975:ESM:360825.360855} algorithm)} \\
      \hline
    \end{tabular}
    \subcaption{Morpho-Lexical features.\label{features:lex}}
  \end{subtable}
  \begin{subtable}[b]{\textwidth}
    \centering
    \begin{tabular}{|ll|}
      \hline
      \multicolumn{2}{|c|}{\textbf{Topics}}\\
      \multicolumn{2}{|p{11cm}|}{10 topics extracted from the Le Monde Corpus}\\
      \hline
      \multicolumn{2}{|p{11cm}|}{Topic features: 10 features, proportion of each topic in each document}\\
      \hline
      \multicolumn{1}{|c|}{Documents}&\multicolumn{1}{c|}{Feature value}\\
      \hline
      \multicolumn{1}{|p{5.5cm}|}{articles containing form}&\multicolumn{1}{p{5cm}|}{max. topic proportion}\\
      \hline
      \multicolumn{1}{|p{5.5cm}|}{concatenation of paragraphs containing form}&\multicolumn{1}{p{5cm}|}{topic proportion}\\
      \hline
      \hline
      \multicolumn{2}{|p{11cm}|}{Newspaper: The newspaper(s) the form appeared in.}\\
      \hline
    \end{tabular}
    \subcaption{Contextual or thematic features.\label{features:struct}}
  \end{subtable}
  \caption{Overview of features used in the classification task.\label{tab:features}}%
\end{table}

Since one of the goals of the \textit{Logoscope} project is to provide means for observing the creation of new words in an enlarged textual context we also explore the influence of thematic features on the automatic identification of the new words (Table~\ref{features:struct}). Our hypothesis is that these features supply interesting additional information not provided by form related and morpho-lexical features. 
A first obvious feature describing the context is the \textit{Newspaper} feature (the newspaper where the unknown form was found).
In addition, we attempt to capture the thematic context of the text containing the unknown form using a technique called \textit{topic modeling} \cite{topicmodels}. Topic models (cf. Section~\ref{sec:topic-modeling}) are based on the idea that documents are mixtures of topics, where a topic is a probability distribution over words. Given a corpus of documents, standard statistical techniques are used to invert this process and infer the topics (in terms of lists of words) that were responsible for generating this particular collection of documents. The learned topic model can then be applied to an unseen document and we can thus estimate the thematic content of this document in terms of the inferred topics. 

In our experimental setting we use topic modeling as follows. We first assemble a set of \textit{general journalistic themes} from a large collection of newspaper articles. Based on these topics we then estimate the thematic content of the larger textual context of the unknown words we investigate.
Several studies (\cite{blei_topic_2009,hoffman_online_2010,hoffman_stochastic_2013}) show that in general tens or hundreds of thousands of documents are needed for a thorough thematic analysis of this kind and that the number of extracted topics is between 100 and 300. In our preliminary experiment we collected 4,755 articles from the newspapers shown in Table~\ref{features:struct} and restricted the number of extracted topics to 10.

The features for each unknown word are obtained by applying topic modeling to two types of context. The first is obtained by concatenating all the sentences containing the unknown form. We expect the result of the topic analysis on this concatenated context to represent the weight of each topic in the closer phrasal context of the unknown word.
The second type of context we use are the articles containing the unknown forms. We apply the topic analysis to each article containing the unknown word and associate the unknown word with an average of each of the topic proportions over the articles. We expect these features to represent the predominant topics of the articles containing the unknown forms. 


\paragraph{Classification method.} Our new word detection problem is casted into a supervised classification problem in the most straightforward way: the validated new words are considered as positive examples and the remaining unknown forms as the negative examples in the training data. We accounted for the strongly imbalanced data\footnote{81 positive examples vs. 611 negative examples} by oversampling the positive class\footnote{Oversampling is a classification technique which helps to deal with imbalanced data. The instances of the minority class are duplicated in order to obtain approximately as many instances as in the majority class. We used the Weka\cite{hall_weka_2009} cost sensitive classifiers to achieve this.}. We then used the SVM classifier as implemented in LibSVM \cite{LibSVM2011}\footnote{We used an exponential kernel with cost 1 and $\gamma=0$} to perform the classification. 
More specifically, we produced 7 classifications (see Table~\ref{tab:classif-results}), one for each combination of the three groups of features described earlier: the \textit{formal}, \textit{morpho-lexical} and \textit{thematic} feature groups.
We then evaluated these classifications in 10-fold cross-validation by looking at precision, recall and F-measure for the positive class and on the average over both the positive and negative class. In addition we also report the number of validated new words (the true positives).

\paragraph{The results} are shown in Table~\ref{tab:classif-results}.
\begin{table}
  \centering
  \begin{tabular}{|l||rrrr||rrrr|}
    \hline
    \multicolumn{1}{|c}{} & \multicolumn{4}{c||}{\emph{formal, morpho-lex, thematic}} & \multicolumn{4}{c|}{\emph{formal, morpho-lex}} \\
    \hline
    \textbf{class} & \multicolumn{1}{c}{\textbf{Prec}} & \multicolumn{1}{|c}{\textbf{Rec}} &  \multicolumn{1}{|c|}{\textbf{F}} & \textbf{corr.} & \multicolumn{1}{|c}{\textbf{Prec}} & \multicolumn{1}{|c}{\textbf{Rec}} & \multicolumn{1}{|c|}{\textbf{F}} & \textbf{corr.} \\
    \hline
    pos & 0.181 & 0.827 & 0.297 &  & \textbf{0.192} & 0.778 & 0.308 &  \\
    both & 0.868 & 0.548 & \textcolor{orange}{0.625} & \textcolor{orange}{67} & 0.864 & 0.597 & 0.669 & 63 \\
    \hline\hline
    \multicolumn{1}{|c}{} & \multicolumn{4}{c||}{\emph{formal, thematic}} & \multicolumn{4}{c|}{\emph{formal}} \\
    \hline
    \textbf{class} & \multicolumn{1}{c}{\textbf{Prec}} & \multicolumn{1}{|c}{\textbf{Rec}} &  \multicolumn{1}{|c|}{\textbf{F}} & \textbf{corr.} & \multicolumn{1}{|c}{\textbf{Prec}} & \multicolumn{1}{|c}{\textbf{Rec}} & \multicolumn{1}{|c|}{\textbf{F}} & \textbf{corr.} \\
    \hline
    pos & 0.160 & 0.531 & \textbf{0.346} &  &  0.190 & 0.481 & 0.273 &  \\
    both & 0.826 & 0.625 & 0.693 & 43 & 0.832 & 0.704 & \textcolor{green}{\textbf{0.752}} & \textcolor{blue}{39} \\
    \hline\hline
    \multicolumn{1}{|c}{} & \multicolumn{4}{c||}{\emph{morpho-lex}} & \multicolumn{4}{c|}{\emph{thematic}} \\
    \hline
    \textbf{class} & \multicolumn{1}{c}{\textbf{Prec}} & \multicolumn{1}{|c}{\textbf{Rec}} &  \multicolumn{1}{|c|}{\textbf{F}} & \textbf{corr.} & \multicolumn{1}{|c}{\textbf{Prec}} & \multicolumn{1}{|c}{\textbf{Rec}} & \multicolumn{1}{|c|}{\textbf{F}} & \textbf{corr.} \\
    \hline
    pos &  0.132 & 0.827 & 0.227 &  &  0.129 & \textbf{0.889} & 0.225 &  \\
    both & 0.836 & 0.350 & 0.415 & 67 & 0.844 & 0.295 & \textcolor{blue}{0.338} & \textcolor{green}{\textbf{72}} \\
    \hline
  \end{tabular}
  \begin{tabular}{|l||rrrr|}
    \hline
    \multicolumn{5}{|c|}{\emph{morpho-lex, thematic}} \\
    \hline
    \textbf{class} & \multicolumn{1}{c}{\textbf{Prec}} & \multicolumn{1}{|c}{\textbf{Rec}} &  \multicolumn{1}{|c|}{\textbf{F}} & \multicolumn{1}{c|}{\textbf{corr.}}\\
    \hline
    pos & 0.136 & 0.877 & 0.236  & \\
    both & 0.851 & 0.345 & 0.404 & 71 \\
    \hline
  \end{tabular}
  \caption{Classification results. In \textcolor{green}{green} (\textcolor{blue}{blue}) best respectively worst F-measure and true positive results. Best balanced results highlighted in \textcolor{orange}{orange}. Overall best precision, recall and F-measure results are set in bold face.}
  \label{tab:classif-results}
\end{table}
The best F-measure (highlighted in \textcolor{green}{green}) for the global classification task was obtained using the \textit{formal} features, but in this setting the  smallest number of validated new words could be identified (in \textcolor{blue}{blue}). 
The highest number of validated new words could be identified using the \textit{thematic} set of features but in this case the global F-measure was comparably low. The best balance between global F-measure and detected new words was obtained using the \textit{formal, morpho-lex, thematic} feature combination.
Overall the results show that using the machine learning techniques presented here, the unknown words filtered by our system can be reordered and presented to a human expert in a more meaningful way. Table~\ref{tab:lex-unknown-word-list} shows a possible reordering produced by our system. 
\begin{table}
  \centering
  \begin{tabular}{ccc}
    ultra-présent ($-$) & crypto-fascisme ($-$) & anti-défilé ($-$) \\
    Etat-département ($-$) & semi-itinérants ($-$) &  pro-MDC ($-$) \\
    anti-alcoolisme ($-$) & mini-Internationale ($-$) & anti-monégasque ($-$) \\ 
    pagano-satanisme (\textcolor{red}{$+$}) & neo-retraité (\textcolor{red}{$+$}) & entraîneur-athlète ($-$) \\
    watts-étalons ($-$) & écarts-types ($-$) & néonicotinoides ($-$) \\
    auto-diagnostiqués ($-$) & agroécologiste (\textcolor{red}{$+$}) & \ldots \\
  \end{tabular}
  \caption{Unknown words ranked by SVM probability. Classification obtained with \textit{form, lex, theme} features. In parentheses: if validated (\textcolor{red}{$+$}) or not.}
  \label{tab:lex-unknown-word-list}
\end{table}
They also suggest that the features used in our experiments were sufficiently powerful to support this classification scenario outweighing the unbalanced data and the difficulty of the classification task.
\paragraph{Groups of features.}
The thematic features played an important role in the classification, since the classification model based on combinations involving the thematic features achieved good results. Thus the classification using the thematic features had the best recall score for the positive class and the F-score for the positive class was highest for the combination of thematic and formal features. 
First this confirms our intuition that the context is helpful at the detection of new words, a finding in line with an important line of work in (Textual) Linguistics where word creation is found to correlate with certain discourse types and textual genres (see Section~\ref{sec:project-background}). Second, this highlights the benefit of using topic modeling for an additional representation of thematic content. 
Indeed, this way it is possible to assess the semantic content of a wider textual context than the more limited co-occurence windows which are used ordinarily.
This aspect, as shown in Section~\ref{sec:previous-work} is rarely taken into account, theoretically or practically, by recent neologism detection utilities.
\paragraph{Qualitative discussion.} For a qualitative analysis 
we applied the classification models based on the \textit{formal}, \textit{morpho-lex}, \textit{thematic} and \textit{formal, morpho-lex, thematic} feature groups on our data and examined the number of correctly identified new words and, for each group, the five best scoring unknown words and the five best scoring validated new words (Table~\ref{tab:predictions}). This allows to better tease out the effect of the various types of features on the selection of valid new words.

First we observe that the \textit{formal} features help identify particularly long new words, and those containing a hyphen (first line). 

The second line shows that the words scoring best with the \textit{morpho-lex} features are mainly compositions (with or without hyphen) or contain a prefix (\textit{anti, non}). 

With respect to the contextual (\textit{thematic}) features we observe on the positive side that they permit the detection of new words with no prominent property (\textit{agnélise, retricoté}), but on the negative side these thematic features seem to favour the selection of words which are not plausible considering traditional formation rules for French word forms (\textit{e.g. schlopp, gesagt}). A closer look at the new words detected through the \textit{thematic} features but not via \textit{morpho-lex} and \textit{formal} features confirmed their ability to select new words with less characteristic forms. Thus, some new words identified by the \textit{thematic} classifier, but not by the \textit{morpho-lex} and \textit{formal} classifiers are: \textit{accrobranches, caricatureurs, conflicté, frenemies, instinctivores \ldots}.
\begin{table*}
  \centering
  \begin{tabular}{|c|c|p{3.5cm}|p{3.5cm}|}
    \hline
    \textbf{Features} & \textbf{\#neos} & \textbf{top 5 valid new words} & \textbf{top 5 (new word?)}\\
    \hline
    \textit{form} & 37 & supermédiateur, doublevédoublevédoublevé, auto-diagnostiqués, néo-célibataires, sur-monétisation & styliste-couturière (no), E-DÉTOURNEMENTS (yes), supermédiateur (yes), garde-à (no), doublevédoublevédoublevé (yes) \\
    \hline
    \textit{lex} & 48 & agroécologiste, multiactivité, auto-obscurcissant, neo-retraité, macrostabilité & agroécologiste (yes), anti-alcoolisme (no), anti-salazariste (no), non-audition (no), multiactivité (yes)\\ 
    \hline
    \textit{theme} & 48 & e-détournements, partenadversaires, hollandisme, retricoté, agnélise & tuitte (no), e-détournements (yes), schlopp (no), gesagt (no), schloppa (no) \\
    \hline
    \textit{form-lex-theme} & 60 & pagano-satanisme, auto-diagnostiqués, neo-retraité, agroécologiste, e-détournements & ultra-présent (no), Etat-départements (no), anti-alcoolisme (no), pagano-satanisme (yes), watts-étalons (no)\\ 
    \hline
  \end{tabular}
  \caption{Top 5 predictions when applying the model.}
  \label{tab:predictions}
\end{table*}

In the current version of the \textit{Logoscope}, the unknown words are ranked using an SVM classifier based on the \textit{formal, morpho-lex, thematic} feature combination. Since this study confirmed the importance of the thematic features we produced a more enhanced topic model of 100 topics based on the Le Monde corpus and use this model to compute the thematic features. This topic model is also used at the assessment of the thematic context of the new words (and in consequence at their documentation) and will be described more in detail in the next section.

\subsection{Thematic Analysis}
\label{sec:thematic-analysis}

The goal of our thematic analysis is to give a general idea about the thematic context in which word creation occurs. We illustrate this goal by an example produced by the \textit{Logoscope} and shown in Figure~\ref{fig:ThemAnal-ex}. The presented article gives an account of several statements issued by the Pope at a Christmas celebration.
The thematic analysis found that the text was mainly about a societal topic and showed that it also deals with some issues related to the theme of defense in an international context. We consider that this gives a reasonably good idea about the general thematic background of the subject addressed in the text, even though it is far from describing the exact content.

Ideally for this type of analysis one would need, in the first place, a register as complete as possible, of journalistic topics or themes (\textit{e.g. Société/Society} or \textit{Défense/Defence} in the example). Unfortunately, to our knowledge such a register is not available. Typically newspaper outlets do assign their articles to particular categories, but the resulting categorisation is neither systematic, nor uniform and can not be expected to be complete. We therefore first developed a method to obtain such a register of general journalistic topics, reflecting the themes a newspaper article might be about. 

Based on this set of journalistic topics, we then developed a method to automatically associate a newspaper article to those topics which best reflect its thematic content. The result is a thematic analysis as shown in Figure~\ref{fig:ThemAnal-ex}. Later, in \textit{Logoscope}'s documentation stage, the topics/themes which were found to be prevalent for that article are associated to the article, and by proxy to the new words it contains.

In the following we first describe the acquisition of the general journalistic topics and then our method to assess the thematic content of a newspaper article.

\paragraph{Acquiring general journalistic topics.}
As already discussed in Section~\ref{sec:previous-work}, one of the few eligible methods to automatically represent and assess thematic content is topic modeling (\cite{blei_topic_2009}). In a nutshell, in topic modeling, documents are viewed as a mixture of topics, where a topic is a probability distribution over words. Standard statistical techniques allow to invert this process and infer the topics (in terms of lists of words) that were responsible for generating the given collection of documents. The learned topic model can then be applied to an unseen document and the thematic content of this document can thus be estimated in terms of the inferred topics.

We acquired the topics from the Le Monde corpus, a collection of more than 900~000 newspaper articles from the French journal \textit{Le Monde}, dating from 1987 to 2006\footnote{We used the gensim toolkit to compute the topic model (\cite{gensim}, \url{https://radimrehurek.com/gensim/}).}. Because of the nature of this corpus, we expect the resulting topic model to reflect the content of the newspaper articles collected by the \textit{Logoscope} fairly well. To this corpus we applied Latent Dirichlet Allocation (LDA), a probabilistic graphical model and generated a set of 100 general journalistic topics. These topics are represented as probability distributions over words. Figure~\ref{fig:ex-topics} shows the words with highest probability for some of these topics. We found that many of these topics were meaningful, in the sense that it seemed doable to figure out the ``latent'' theme they represent and give it a suggestive label. For example, one could say that Topic 0 in Figure~\ref{fig:topic-0} is about economics and Topic 2 in Figure~\ref{fig:topic-2} about law or justice. For others it was obvious that the underlying theme was not interesting. An example for this is Topic 19 in Figure~\ref{fig:topic-19} which apparently is mainly a collection of male first names.\footnote{To our knowledge currently there is no principled way to automatically ``label'' the topic lists produced by this method.}

\begin{figure}
  \centering
  \begin{subfigure}[b]{0.31\textwidth}
    \begin{tabular}{ll}
      groupe & 0.05867\\
      société & 0.02616\\
      capital & 0.01642\\
      actionnaire & 0.01381\\
      filiale & 0.01078\\
      affaires & 0.00926\\
      entreprise & 0.00912\\
      franc & 0.00907\\
      milliard & 0.00860\\
    \end{tabular}
    \subcaption{Topic 0\label{fig:topic-0}}
  \end{subfigure}
  \begin{subfigure}[b]{0.31\textwidth}
    \begin{tabular}{ll}
      droit & 0.02779\\
      conseil & 0.02254\\
      loi & 0.01790\\
      pouvoir & 0.01520\\
      commission & 0.01475\\
      rapport & 0.00992\\
      cas & 0.00952\\
      devoir & 0.00805\\
      texte & 0.00794\\
    \end{tabular}
    \subcaption{Topic 2\label{fig:topic-2}}
  \end{subfigure}
  \begin{subfigure}[b]{0.31\textwidth}
    \begin{tabular}{ll}
      jean & 0.23761\\
      pierre & 0.09342\\
      claude & 0.05313\\
      jacques & 0.05284\\
      marier & 0.05010\\
      louis & 0.03668\\
      alain & 0.03400\\
      marc & 0.02243\\
      saint & 0.01784
    \end{tabular}
    \subcaption{Topic 19\label{fig:topic-19}}
  \end{subfigure}
  \caption{Words with highest probabilities in some topics acquired by applying Latent Dirichlet Allocation to the Le Monde corpus. Topic 0 and Topic 2 are examples of meaningful topics in our setting, which could be labeled easily. In contrast Topic 19 shows a topic which could hardly be used for representing the content of newspaper articles.\label{fig:ex-topics}}
\end{figure}

Hence we could not use the topics directly but had to remodel them. For this we selected the most interesting topics (currently 71). In these topics we only kept the significant terms and also added some terms if necessary. This had to be done manually and required a considerable effort. It resulted in a collection of 71 general journalistic themes which are meaningful to human readers and reflect the content of the journals we are dealing with.

Topic models generated by LDA only can easily be used in a subsequent step to infer, for an unseen text, a probability distribution of the learned topics. Because in the \textit{Logoscope} we modified the words in the topics, this computationally cheap automatic inference is no longer available.

Figure~\ref{fig:ex-themes} shows two examples of resulting themes.\footnote{It also shows that the terms in the themes comprise the POS (grammatical category) and that we also use compound terms, in an effort to increase accuracy and avoid lexical ambiguity. 
 We obtained the grammatical categories by preprocessing the texts with TreeTagger and adding the corresponding grammatical categories to the terms in the themes. The themes also contain compound terms, which are often highly characteristic of a particular topic. Thanks to lemmatisation and POS tagging it is possible and technically not to expensive to match them in the articles.} In contrast to the topics, we also no longer dispose of the probability weights and therefore all terms are equally important for the thematic analysis of a text.

\begin{figure}
  \centering
  \begin{subfigure}[b]{0.49\textwidth}
    \begin{tabular}{l}
eurogroupe-nc\\
société-nc général-adj\\
capital-nc\\
capitalisme-nc\\
capitaliste-adj\\
capitalistique-adj\\
actionnaire-adj\\
actionnaire-nc\\
actionnarial-adj\\
    \end{tabular}
    \subcaption{Economie-Finance (economy and finance).\label{fig:eco-fin}}
  \end{subfigure}
  \begin{subfigure}[b]{0.49\textwidth}
    \begin{tabular}{l}
droit-nc\\
avoir-v droit-nc\\
chambre-nc du-prp conseil-nc\\
conseils-nc\\
loi-nc\\
loi-nc carrez-nc\\
pouvoir-nc\\
commission-nc\\
commission-nc européen-adj
    \end{tabular}
    \subcaption{Droit-JusticeLegislation (law, justice, legislation)\label{fig:droit}}
  \end{subfigure}
  \caption{Examples of resulting themes: Economie-Finance (economy and finance) and Droit-JusticeLegislation (law, justice, legislation).\label{fig:ex-themes}}
\end{figure}

\paragraph{Assessing thematic content.}
Based on the themes developed this way we implemented our thematic analysis as follows. Input to our method are the 71 general journalistic themes and a text, typically a newspaper article. The result is on one hand a thematic colouring of the text as shown in Figure~\ref{fig:ThemAnal-ex}. On the other hand the text is also associated with the most relevant themes -- this is what we presume the text is about\footnote{In the example in Figure~\ref{fig:ThemAnal-ex} these are \textit{Mythologie, Réligion et Croyance} (mythology, religion, religious faith), \textit{Conflits Armés Contemporains} (contemporary armed conflicts) and \textit{Afrique} (Africa).}.

We determine the most relevant themes by looking at the terms the themes and the text have in common. The more terms shared by a theme and the text, the more relevant we consider the theme for this text\footnote{To compute the term overlap efficiently we use again the \citealt{Aho:1975:ESM:360825.360855} algorithqm.}. 

To wrap up, with this flavour of thematic analysis we found a viable way to assess the thematic context of new words and to associate them with labeled themes which are more meaningful for human readers than the topics resulting from topic modeling.

\section[Applications]{The \textit{Logoscope} at Work: Linguistic and Extra-linguistic Applications}
\label{sec:application}

The aim of this section is to give a general idea about how the
\textit{Logoscope} could be used in an analysis of word creation.  We
first present some findings about word creation in French newspaper
texts which we could obtain using the \textit{Logoscope}.  We then
illustrate its use for classical linguistic and lexicographic analyses
but also for some basic observations pertaining to the realm of social
and/or political sciences.

At the time of writing (end of 2017), the \textit{Logoscope} data base contained 1126 new words together with the almost 15 000 paragraphs in which they occured. Most occurences appeared in the journal ``Les Echos'', followed by ``Le Figaro'' and ``Libération''. However, to be more meaningful, these numbers still need to be related to the total number of paragraphs collected for each journal.

In almost 60\% of instances the new words were used as common nouns and the second most frequent category were adjectives (almost 20\%). Interestingly, almost 18\% of instances were used as proper nouns. Some examples are various names used to write about newly built administrative units, as \textit{alsace-champagne-ardenne-lorraine} or \textit{haut-de-france} (see below for some more discussions). Others are derived from proper names by affixation (\textit{dieselgate, ex-erdf}) but there are also genuine innovations as for example \textit{Boxit}, a contraction of \textit{Boris} and \textit{exit}. 

Regarding the creation process, most new words (almost 60\%) emerged from a morpho-semantic process and almost 40\% are also loan words. Some sample new words which were created by borrowing and/or a morpho-semantic process are shown in the following.

\begin{enumerate}
\item\label{item:1} Pour utiliser une monnaie virtuelle, il faut disposer d'un portefeuille électronique, qui stocke bitcoins, \textbf{ethers} ou ripples. Les portefeuilles sont disponibles sur les smartphones via des applications ou sur les ordinateurs via des sites dédiés.
\item\label{item:2} Lycéenne le jour, \textbf{startuppeuse} la nuit. A 17 ans, Philippine Dolbeau développe New School depuis déjà un an et demi. Ce système vise à remplacer et à automatiser les cahiers d' appel à l'école. Lauréate de plusieurs prix, elle est notamment suivie par Apple.
\item\label{item:3} Il existe des « professeurs », des « docteurs » et des « restaurateurs », mais pourquoi pas des « \textbf{professeuses} », « docteuses » ou « restaurateuses » ? Claude\label{item:4} Duneton (1935-2012) analysait l'impossible féminisation des mots en « -eur ». Le Figaro vous fait redécouvrir sa chronique.
\item Va-t-on assister à la naissance d'un \textbf{assado-poutinisme}, ultime avatar d'une politique étrangère décidément incapable de renouer avec la mesure et l'intelligence de notre nouveau monde ? (\ldots)
\end{enumerate}

In the first example (\ref{item:1}) the new word \textit{ethers} is a pure loan word whereas in the second example (\ref{item:2}), \textit{startuppeuse} the borrowing is combined with a morpho-semantic process (affixation). Finally the last two examples (\ref{item:3}, \ref{item:4}) show new words (\textit{professeuses}, \textit{assado-poutinisme}) which emerged through a morpho-semantic process only (affixation and composition resp.). Example~\ref{item:3} is also interesting from a lexicographic point of view because it points to the fact that in French there are few feminisations of words in ``-eur'', therefore those which are created are particularly outstanding.
These examples show how the \textit{Logoscope} could be used for some traditional lexicographic investigations, as for example observing which are the most productive word-creation processes, or which affixes are most frequently used in the word creation process.

The position where the new words appear gives some hints about some journalistic practices.
In most cases (54\%) they appear in the middle of the newspaper articles, which is not surprising, since this is the main body of the texts. However an important portion (30\%) of the instances are used in the last three paragraphs, a relatively small part of the article, suggesting that a common praxis was to ``drive home'' a point by using a surprising new word. In contrast, new words were relatively rarely used in introductory paragraphs (16.5\%)\footnote{The introductory paragraphs include the title and chapeau because it was not possible to distinguish them from the main body paragraphs.} -- apparently they were not found to have a sufficiently attention-grabbing effect.

\subsection{New Words and Textuality: going deeper in the study of lexical creativity}
\label{sec:textuality}

The \textit{Logoscope} naturally allows for a traditional
lexicographic analysis of new words. But due to its specificity of
taking into account the textual context of word creation it also
permits to study these new words there where they are used: in the
surrounding texts. 
For example, as our brief discussion above shows, it allows empirical observations regarding
journalistic style. 

However, while other systems exist which allow to observe the
interaction of word creation with journalistic style, the
\textit{Logoscope} also makes possible more systematic empirical
studies at a larger scale due to the integrated more extended
contextual documentation.  More specifically, to our knowledge there
are no other new word detection systems which collect data about the
thematic context where the word creation occurred.

The following very sketchy analysis of word creation in terms of its
larger thematic environment is meant to 
give an idea of how this specificity of the \textit{Logoscope} could be used.

\begin{figure}
  \centering
  \begin{subfigure}[b]{0.49\textwidth}
  \includegraphics[width=\textwidth]{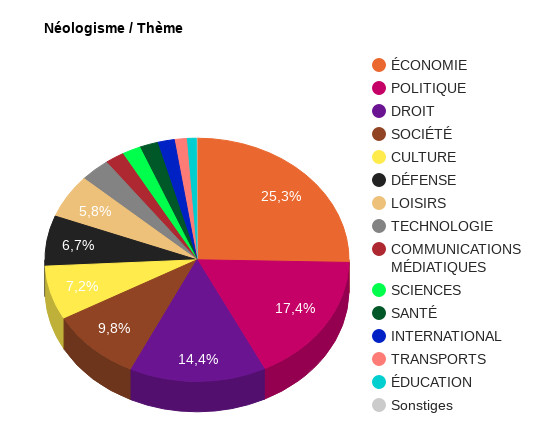}
    \subcaption{Topics of newspaper articles where most new words were used.\label{fig:neo-in-themes}}
  \end{subfigure}
  \begin{subfigure}[b]{0.49\textwidth}
  \includegraphics[width=\textwidth]{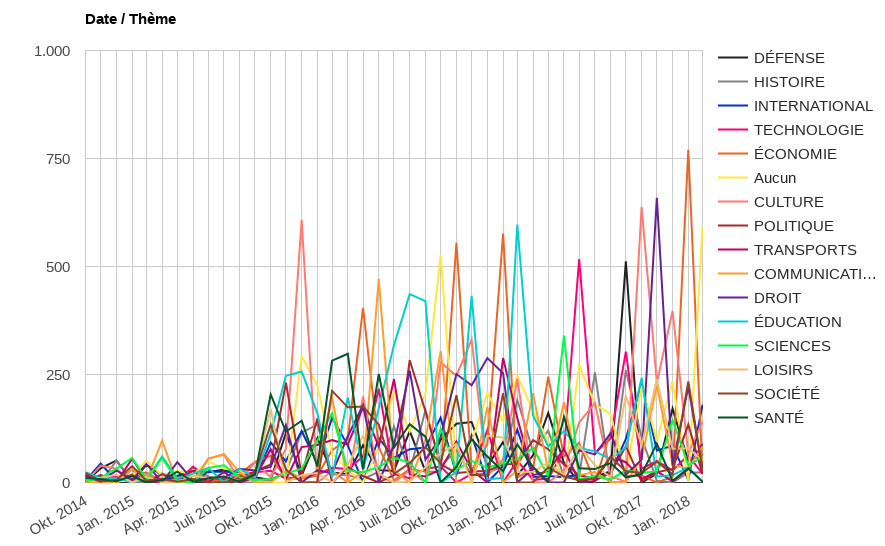}
    \subcaption{Occurrences of new words in articles about different topics over time.\label{fig:date-theme}}
  \end{subfigure}
  \caption{Observing the thematic context in which word creation occurs. The figures are screenshots obtained from the online version of the \textit{Logoscope} (\protect\url{http://logoscope.unistra.fr/topic/}).}
\end{figure}

First, Figure~\ref{fig:neo-in-themes} gives an overview of the topics of the articles where most new words were used. Even though to be more meaningfull this should be related to the overall distribution of articles over the topics, we think that this visualisation nevertheless allows a first rough estimation of the thematic context where most new words were used. We see that new words mostly appeared in articles dealing with economics (25\%) followed by articles about politics (17\%) and law (14\%). Figure~\ref{fig:date-theme} shows in what thematic contexts most new words were used over time. Thus in January 2016 most collected new words were used in articles with a cultural background, whereas in February 2017 the articles containing the new words were mostly related to culture. ``Zooming into'' the time frame of January 2016 we can find some examples of new words used in a cultural thematic context:\footnote{New words are shown in bold-face, we don't show the entire paragraph.}

\begin{itemize}
\item Oui, beaucoup ont espéré qu'on se fasse tuer. \textbf{tu-er} », poursuit-il, en rappelant la fragilité du journal (\ldots)\footnote{This is about the ``Charlie Hebdo'' newspaper}
\item \ldots le FFZero1 concept apparaît comme un \textbf{hypercar} très inspiré de la Batmobile de Batman \ldots
\end{itemize}

Finally we give some examples of new words used in articles related (among others) to education, from February 2017:
\begin{itemize}
\item En matière de cibles publicitaires, les \textbf{millennials} (20-30 ans) ont fait dernièrement beaucoup parler d'eux.
\item Michel Agier : « Repenser l' engagement citoyen est le seul moyen d'éviter l'\textbf{encampement} des migrants »
\item David Forge, jeune agriculteur et \textbf{youtubeur} le 30 Janvier 2017 à Saint Senoch .
\end{itemize}

\subsection{Spotting and Documenting New Realia in French Society (2014-2016)}
\label{sec:new-realia}

In this last section, we would like to briefly show how the \textit{Logoscope}
contributes, beyond lexicologic and lexicographic issues
(traditionally associated with new words), to spot and document the
emergence of new French realia, that is realities which are specific
to contemporary French society.

Except for the domain of sociolinguistics, psycholinguistics and
discourse studies (e.g. \cite{dufour_rosier}), the approaches adopted
by linguists often implement a ``language for language's sake''
perspective, which prevents taking into consideration 
extra-linguistic realities. Concerning new words, this word-only
oriented perspective is evident in a large scale of studies, ranging
from the design of language specific rules (e.g. \cite{corbin}) to the
description of textual phenomena
(\cite{gerard:halshs-01093209,gerard:halshs-01093234}).
This approach is certainly justified by the fact 
that most of the many new words
created each year originate in a rhetorical intention (to denounce, to
joke, to emphasise, etc.) rather than in a need to communicate about
an extra-linguistic reality (technical, political, economical,
environmental, etc.). The consequence is however that these extra-linguistic realities are not observed and studied.

With respect to such extra-linguistic realities
 new words can be classified in two groups according to their ``cultural spread''.
\begin{enumerate}
\item Words which have been created within a specific cultural area. They reflect realities which, at the time we experience them, are only happening in this area (\textit{i.e.} realia). We assume therefore that they do not concern any other cultural area in the world.
\item Words refering to new realities which, mostly because of globalisation, also appear in other cultural areas.
\end{enumerate}
Table~\ref{tab:realia} shows some examples from different languages for these two groups of new words.
\begin{table}
  \centering
  \begin{tabular}{|l|l|l|}
    \hline
    \multicolumn{1}{|c}{\textbf{Language}} & \multicolumn{1}{|c}{\textbf{Non-cultural specific}} & \multicolumn{1}{|c|}{\textbf{Cultural specific}} \\    
    \hline
    English & vlogging, retweet, brexit & Mx., beer o'clock \\
    Italian & svapare, macaron, jihadista & enopirateria, sciarpata \\
    Catalan & bitcoin, crowdfunding, wok & iaioflauta, xirucaire \\
    French  & téléverser, Zika, nomophobie & panthéonisable, CRDS \\
    \hline
  \end{tabular}
  \caption{Examples of cultural specific and non-cultural specific new words added in dictionaries between 2014 and 2016. }
  \label{tab:realia}
\end{table}
Since its launch in October 2014, the \textit{Logoscope} is spotting new words designating emerging cultural specific realities. For example, in this period of time, a notable new word showing increasing frequency during several months was \textit{zadiste}. 

\textit{Zadiste} is derived from ``ZAD'' which is an acronym for ``zone à défendre'' (area to be protected). 
The \textit{Logoscope} data showed that after its appearance its frequency in the newspaper articles was relatively stable during several months. The thematic analysis revealed that it mostly appeared in textual contexts related to social conflicts. Thus the tool not only documented the appearance of this new word but also showed how its propagation went along with the establishment and consolidation of a new social conflict.

Other examples of how lexical creation testifies of political events are the new words \textit{dealmaker} which was first detected in November 2016 when Donald Trump was elected president of the United States, and \textit{Bruxellistan} which appeared between January and March 2016 when a suspect from the Paris terror attacs was arrested several months later in Brussels. Here the word \textit{Bruxellistan} arguably also shows a rhetorical intention, namely to denounce a situation in Brussels and to put it on a level with a similar situation in London (which at some point was called \textit{Londonistan}).

\begin{figure}
  \centering
    \begin{subfigure}[b]{0.49\textwidth}
\includegraphics[width=\textwidth]{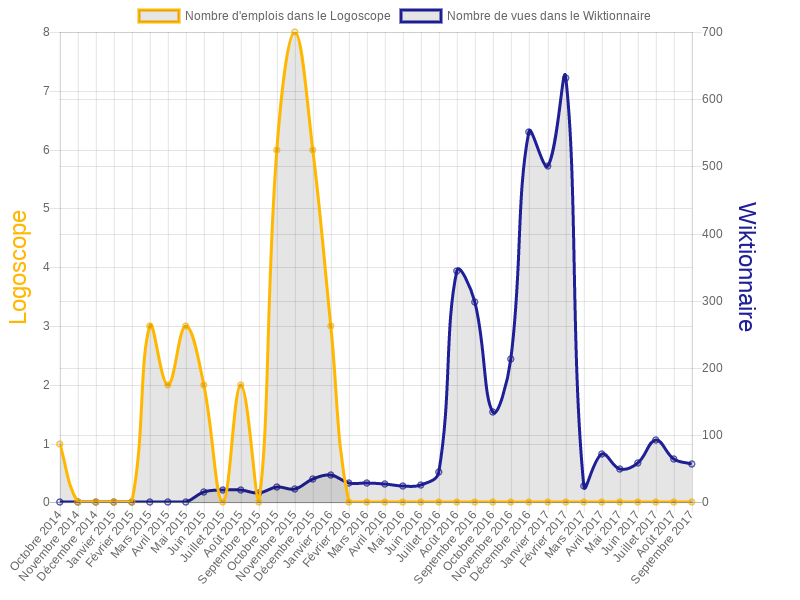}
\subcaption{\textit{bitcoin}}
    \end{subfigure}
    \begin{subfigure}[b]{0.49\textwidth}
\includegraphics[width=\textwidth]{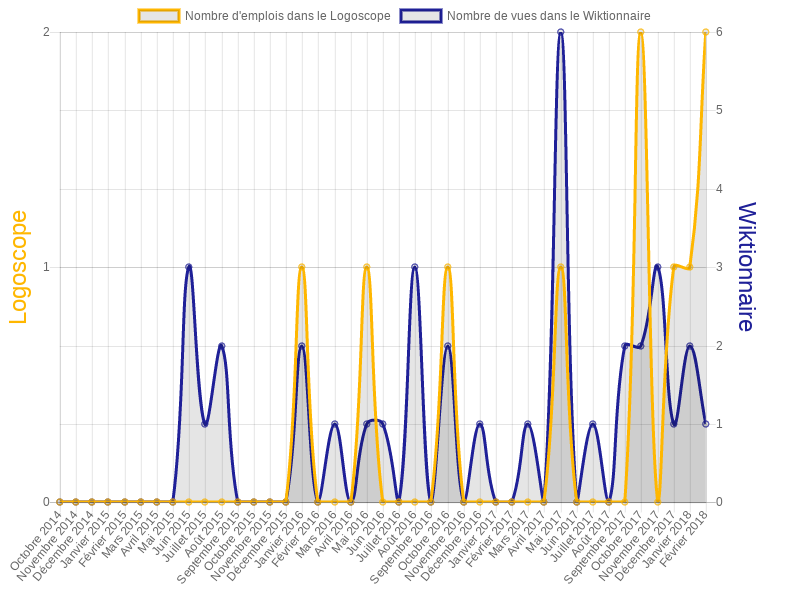}
\subcaption{\textit{ethers}}
    \end{subfigure}
  \caption{Occurences of new words \textit{bitcoin} and \textit{ethers} in the \textit{Logoscope} vs. lookups in the Wiktionary.}
  \label{fig:bitcoin-ethers}
\end{figure}

A word creation accompanying  an economic event is \textit{bitcoin}. Figure~\ref{fig:bitcoin-ethers} shows when it was observed by the \textit{Logoscope} (yellow lines) compared to when it was looked up in the Wiktionary (blue lines).\footnote{\url{https://fr.wiktionary.org/}} According to this it first appeared in the (monitored) press before it was available in the Wiktionary. In contrast, \textit{ethers}, a word creation related to the same economic event was visible in the Wiktionary first. 

We also found that word creations ending in \textit{-gate} were reliable indicators for various scandals: \textit{volkswagengate} and \textit{dieselgate} accompanied the diesel emission scandals and \textit{fillongate} appeared when a French presidential candidate was suspected of corruption.

Another interesting new word detected was \textit{Hauts-de-France} which designates the name of a newly created administrative unit (a region) in France. Monitoring this word with the \textit{Logoscope} allowed to testify of continuity and disruptions in this process of geopolitical change. Another administrative unit created at this point in France was the region which later was called \textit{Grand-Est}. Whereas obviously \textit{Hauts-de-France} was used by journalists, \textit{Grand-Est} did not appear in the \textit{Logoscope} collection, suggesting that this territory was less perceived as a genuine traditional or historical entity than the \textit{Hauts-de-France}. Instead, we found several other denominations as for \textit{e.g.} \textit{alsace-champagne-ardenne-lorraine}, \textit{alsace-lorraine-champagne-ardenne} or \textit{alsace-champagne-ardenne}, which are all compositions made up of the names of the original regions.

Other word creations, like \textit{bio-sourcé, bio-morale, bio-éthiciens}, are built using a particular prefix (in this case \textit{bio} (organic)) which largely determines the new word's meaning. In the case of \textit{bio} it also arguably indicates the development of a new social identity centered around an organic style of living. 

Some other word creations arguably point to phenomena related to the perception of gender identity. 
Thus, many new words designating occupations (or occupational titles) appear with either a female morphology only or both female and male morphology. Some examples are \textit{clippeuse, proffesseuse, startupeuse, youtoubeuse}. 

These examples show that this data and way of monitoring lexical change may be of direct interest not only for lexicographers and linguists but also for journalists, historians and researchers in social and political sciences.

\section{Conclusion}
\label{sec:perspectives}

This paper described the \textit{Logoscope} framework -- a tool for
the detection and documentation of French new words in online
journalistic publications. 

Compared to other available new word detection systems, the
\textit{Logoscope} belongs to a more elaborate category of frameworks
consisting of a detection phase based on a methodologically well
defined dynamic corpus and a documentation phase where the detected
and documented new words are added to a linguistic knowledge base. In
this paper we described both the corpus and methods used for the
detection and the theoretic linguistic and extra-linguistic principles
guiding the documentation of the new words.

We showed how the \textit{Logoscope} benefits from recent
developments of natural language processing techniques to largely
automate both the detection and the documentation steps and to thereby allow a more
extensive and comprehensive
empirical study of the emergence of new words.

The \textit{Logoscope} is targeted not only at the traditional user
categories of such a tool (\textit{e.g.} linguists and lexicographers)
but more generally at user groups concerned with how various cultural,
societal or political developments are reflected in journalistic
publications (\textit{e.g.} journalists but also social scientists or
economists). We showed how this viewpoint determined the documentation
of new words in the \textit{Logoscope} framework. One consequence was
our choice to deliberately restrict the description of the lexical
features such that the annotation effort is reduced but a subsequent
in depth linguistic or lexicographic analysis remains possible. In
contrast, we opted for more extensive information about contextual
features of the new words, as for example the topics addressed by the
articles containing them. To our knowledge the \textit{Logoscope} is the
only neologism detection system providing this feature automatically.

The \textit{Logoscope} new words base is publicly available and can be queried online
at the address \url{http://logoscope.unistra.fr}.

\bibliographystyle{spbasic}      
\bibliography{Logoscope,Logoscope_zotero}   

\end{document}